\pdfoutput=1

\documentclass[11pt]{article}

\usepackage[preprint]{acl}

\usepackage{times}
\usepackage{latexsym}

\usepackage[T1]{fontenc}

\usepackage[utf8]{inputenc}

\usepackage{microtype}

\usepackage{inconsolata}

\usepackage{tikz}
\usepackage{amsmath}
\usepackage{subcaption}
\usepackage{listings}
\usepackage{graphics}

\usepackage{url}

\usepackage{breakurl}

\usepackage{multirow, makecell}
\usepackage{booktabs}

\newcommand{\remove}[1] {}

\newcommand{\fullpaper}[1] {}

\makeatletter
\lst@Key{countblanklines}{true}[t]%
    {\lstKV@SetIf{#1}\lst@ifcountblanklines}
    
\lst@AddToHook{OnEmptyLine}{%
    \lst@ifnumberblanklines\else%
       \lst@ifcountblanklines\else%
         \advance\c@lstnumber-\@ne\relax%
       \fi%
    \fi}
\makeatother

\title{Is (Selective) Round-To-Nearest Quantization All You Need?}
\author{Alex Kogan \\
  Oracle Labs \\
  Burlington, MA \\
  USA \\
  \texttt{alex.kogan@oracle.com}}

\remove{
\author{Alex Kogan}
\affiliation{%
  \institution{Oracle Labs}
   \city{Burlington, MA}
   \country{USA}
}
\email{alex.kogan@oracle.com}
}

\begin{document}

\maketitle

\begin{abstract}
Quantization became a necessary tool for serving ever-increasing Large Language Models (LLMs).
RTN (Round-to-Nearest) is perhaps the simplest quantization technique that has been around 
well before LLMs surged to the forefront of machine learning (ML) research.
Yet, it has been largely dismissed by recent and more advanced quantization methods that claim superiority over RTN 
in nearly every aspect of performance.
This work aims to dispel this established point of view, showing that RTN is not only much cheaper to apply, 
but also its token generation throughput can be better than and accuracy can be similar to more advanced alternatives.
In particular, we discuss our implementation of RTN based on the recent Marlin kernels and 
demonstrate how the accuracy of RTN can be gradually improved by selectively increasing 
the data precision format of certain model layers and modules.
Based on our results, we argue that RTN presents a viable and practical choice for quantizing LLMs.

\end{abstract}

\section{Introduction}

\remove{
Outline:
-- Inferencing LLMs becomes a critical task with models increasing in size
-- Quantization became de-facto the standard way of serving LLMs
-- Lots of research into quantization techniques, overlooking the RTN
-- Our contribution: improved RTN for better accuracy and speed
}

LLM inference is an expensive process because large models require expensive machines with a large amount of high bandwidth memory (HBM).
In fact, some of the largest models available today, such as Llama3.1-405B~\cite{Llama3.1} or DeepSeek V3/R1 with their 671B parameters~\cite{DeepSeek},
do not even fit into one powerful server with 8 Nvidia H100 GPUs with 80GB HBM each.
Furthermore, LLM inference, especially if it involves a  generative model such as Llama, 
is wasteful because the token generation is a sequential process, producing new tokens one after another.
As such, the computation involved in token generation is comprised of vector by matrix multiplications that do not allow much data to be reused once it is brought from memory to computing units.
Consequently, the generation process is known to be memory-bound~\cite{FCC24, AKP24, KLZ23, KHG24}, leaving all those powerful GPUs underutilized.

Over the past few years, quantization has emerged as de-facto the standard way of tackling both aforementioned issues.
During a quantization process, model weights (and/or activations) are compressed into lower precision (e.g., from 16 bits down to 4 bits), therefore resulting 
in a smaller memory footprint for the model.
This, in turn, leads to reduced hardware requirements for inference and, arguably, to a more efficient token generation process.

Due to its practical importance, quantization of LLMs has attracted lots of interest in the research community.
This lead to a number of impressive results, 
with methods like GPTQ~\cite{FAH22}, AWQ~\cite{LTT24}, SmoothQuant~\cite{XLS23}, QQQ~\cite{ZZH24} and many others, 
each aiming to produce a highly compressed model that is accurate and fast during inference.
Those advanced techniques often dismiss or claim superiority over another simple quantization method that has been around 
well before LLMs came to exist~\cite{Kri18, GKD21}.
This method is RTN or Round-to-Nearest, which involves scaling given model weights and, as its name suggests, rounding the result to the nearest integer.

A few favorable characteristics help RTN to stand out from other, more advanced quantization techniques, 
making it a desirable choice for quantizing LLMs.
First, unlike most other quantization methods, it is \emph{data-free}~\cite{TSW23}, 
i.e., it does not require any calibration data nor a corresponding calibration process.
As such, RTN can be applied on-the-fly (i.e., while loading a model), and its application is 
fast (a few milliseconds or seconds, depending on the size of the model) and cheap.
Second, RTN can be applied to a model even on a machine that 
cannot fit the original (non-quantized) model in its GPU memory.
For instance, we can use a server with 4 Nvidia H100 GPUs to load, quantize with RTN and run inference on the Llama3.1-405B model.
At the same time, RTN is often believed to lag behind more advanced quantization techniques in two crucial areas -- 
accuracy~\cite{FAH22, LTT24, HLQ24} and generation latency.

In this paper, we tackle both those issues. 
First, through extensive evaluation we show that the recovery rate of RTN (i.e., its accuracy across various tasks compared to the baseline model)
\begin{enumerate}
\item is perfect when a model is quantized to 8 bits.
\item is nearly perfect and on-par with other, more sophisticated quantization techniques on an extremely large model quantized to 4 bits.
\item falls behind on moderate and small scale models quantized to 4 bits.
\end{enumerate}
Focusing on the last case, we investigate \emph{selective quantization}, in which certain layers of the Transformer stack of a given model are selected to be quantized
to a different precision, and show the efficiency of this approach in recovering most accuracy loss for RTN.
In particular, we show that leaving just a part of one layer of Llama-3.1 70B model in 8 bit precision, while quantizing the rest of that layer 
and all other 79 layers into 4 bits leads to nearly perfect recovery rate.
Note that this adds less than $0.05$ bits per weight on average, resulting in only insignificant memory increase.

Second, we utilize recent Marlin kernels~\cite{FCC24} (developed originally for GPTQ~\cite{FAH22}) 
to provide highly efficient implementation for RTN.
We integrate those kernels into vLLM~\cite{KLZ23}, a popular framework for serving LLMs, and show that a model quantized with RTN can generate tokens up to $37\%$ 
faster, relative to the baseline model, when the model is quantized into $4$ bits and up to $25\%$ faster when quantized into $8$ bits.
To the best our knowledge, this is the first example of a data-free quantization technique that produces such savings;
moreover, those savings are higher than those achieved with (calibration-data dependent) GPTQ and AWQ, and (data-free) BitsAndBytes~\cite{DLB22}.

\remove{
In this paper, we describe a surprisingly simple way to improve the accuracy of RTN by inverting the scaling factors when appropriate.
Our proposed method, which we call RTN+, preserves all the pros of RTN and does not have any inference time overhead on top of it.
Through extensive evaluation using the a wide selection of tasks from the standard LM-bench suite~\cite{}, we show that RTN+ shrinks, and often eliminates completely,
the accuracy gap from other, more advanced quantization techniques.
To tackle the generation latency inefficiency, we turn to the recently proposed Marlin kernels for mixed precision multiplication~\cite{}.
Those methods power other advanced quantization techniques such as GPTQ and AWQ, and we show that they can provide the same 
state-of-the-art inference performance to RTN+.
Taken together, our work shows that RTN+ presents a viable and practical alternative to recent, more advanced quantization techniques.
}

\section{Related Work}

\subsection{Post-training Quantization}
Unlike quantization-aware training (QAT) strategies that typically come with a high training cost~\cite{HLQ24, ZLL24}, 
post-training quantization (PTQ) offers cheap, accurate and performant methods to reducing the
size of LLMs. 
The majority of PTQ methods require calibration data, which they use to find optimal values, 
i.e., values that reduce a rounding error introduced when decreasing the number of bits representing model weights and/or activations.
For instance, AWQ uses calibration data to find salient weights, which it then scales with factors computed to minimize the 
the mean-squared error of quantization~\cite{LTT24}.
GPTQ uses calibration data to calculate inverse Hessian matrix employed to find quantized weights~\cite{FAH22}.
The QQQ method~\cite{ZZH24} uses calibration data twice, once to identify and selectively smooth activation channels that 
contain outliers, similarly to~\cite{XLS23}, and then to calculate the inverse Hessian matrix similarly to~\cite{FAH22}.

Calibration data-dependent quantization has several drawbacks.
First, as noted by Tang et al.~\cite{TSW23} and demonstrated by Wu et al.~\cite{WXY23}, it might weaken the ability of LLMs to generalize as it introduces the risk 
of overfitting to the data selected for calibration.
Second, it requires a separate process for applying quantization, which results in a separate set of (quantized) model weights.
This makes the application of quantization more cumbersome in practice.
Third, this process, albeit much shorter than model pre-training or even fine-tuning, still requires non-trivial resources, such 
as several hours or even days of GPU time; moreover, it requires at least the number of GPUs with enough HBM to load the original model.

Several papers do offer \emph{data-free} quantization techniques~\cite{XLS23, DLB22, TSW23}.
Their common theme is to separate outliers (weights whose magnitude is significantly large, e.g., more than $6.0$~\cite{DLB22}) 
from other weights; the outliers are then kept in full precision while the rest of the model is quantized either with
uniform (RTN-like)~\cite{DLB22} or non-uniform method~\cite{XLS23, TSW23}.
While the number of outliers is typically small and has a negligible impact on the size of the quantized model, 
their special treatment adds computational overhead, leading to increased token generation latency.
We demonstrate that by including the results of the BitsAndBytes quantization~\cite{DLB22} in our evaluation.

Selective quantization maintains model weights in multiple data formats (e.g., FP16, INT8 and INT4).
As such, it can be considered related to prior work on mixed-precision quantization, e.g.,~\cite{ZZG19, YCR24}.
In fact, some of the aforementioned data-free methods also belong to this group, as they
store outliers and non-outliers in different precision.
Unlike prior work, however, selective quantization does not require model retraining or fine-tuning~\cite{ZZG19} 
nor occasional re-quantization during inference~\cite{YCR24}.
Generally speaking, selective quantization is more coarse-grained and simpler than most prior methods,
which significantly simplifies its integration into an inference framework in practice.

\subsection{RTN}
Round-to-nearest (RTN) quantization is, perhaps, the simplest quantization technique known well before large language models came to exist~\cite{Kri18, GKD21}.
RTN is applied on a vector of real values in a floating point format (e.g., FP16 or BFloat16) by dividing each value in the vector by a scaling factor, 
rounding the result to the nearest integer (hence, the name of this method) and truncating, if necessary, the final integer number so
it fits into the target range of values (e.g., [-128, 127] for INT8 or [-8, 7] for INT4). This method is summarized by the following simple formula:
\begin{equation}
\label{eq:RTN}
RTN(r) = Int(r/S),
\end{equation}
where $r$ is a vector of input values, $S$ is a scaling factor and $Int$ is a function that maps a real value to an integer value through round-to-nearest and truncate operations.
We note that  we describe a \emph{symmetric} variant for RTN; an asymmetric variant, more suitable for input values that are not distributed around zero, 
subtracts an arbitrary integer zero point from the result of the $Int$ function~\cite{GKD21}.

Now we turn our attention to how the scaling factor $S$ is chosen.
The scaling factor essentially maps given real values into a partition in a range of integers, where the size of the range is set by the number of bits in the target precision.
In the symmetric formulation of RTN, $S$ is set according to the following formula:
\begin{equation}
\label{eq:scale}
S = \frac{2max(|r|)}{2^b-1} = \frac{max(|r|)}{2^{b-1}-0.5},
\end{equation}
where $max(|r|)$ calculates the maximum absolute value in the given input and $b$ is the target precision bit width.

As described, RTN is oblivious to the size of the input vector $r$, however, choosing the ``right'' size for the vector has important practical implications.
Large vectors result in quantized models being susceptible to a significant accuracy loss
due to the presence of outliers in the input data~\cite{AZY25, TSW23, DLB22}, which tend to reduce the quantization resolution by skewing the scaling factor.
Hence, the common approach (also used in this paper) splits each weights tensor along its channels and then each channel is split into groups of $g$ values;
RTN is applied separately to each group.
This sub-channel-wise application contains outliers better (now their impact is limited to a group of $g$ numbers rather than the entire channel/tensor), but 
elevates the memory cost as it requires storing more scaling factors.
$g$ is typically set to $2^n$, with $n=7$ being seemingly a popular choice~\cite{FCC24, ZZH24}.
We note that the issue of dealing with outliers in the input is far from being unique to RTN, and in fact, is in the heart of most quantization techniques for LLMs~\cite{XLS23, TSW23, DLB22, ZZH24, LTT24};
likewise, refining the application of quantization by splitting input values into small groups is a common technique for improving accuracy of quantized models~\cite{FAH22, ZZH24}.

\subsection{Marlin kernels}
Marlin kernels~\cite{FCC24} have been developed to support high-performance mixed-precision GEMM (general matrix multiplication) operations.
Such operations allow multiplying 16 bit activations by 4 or 8 bit weights (commonly referred to as W4A16 or W8A16 in the literature~\cite{ZZH24, LTT24}).
A common, but inefficient way of implementing them is by dequantizing the weights first and then applying GEMM in 16bit precision.
Marlin hides the overhead of dequantization by effectively overlapping the data access latency with floating-point operations 
involved in dequantization and GEMM calculations.
It does so through the combination of various techniques, such as an asynchronous memory access and optimized data layout in memory, 
According to Frantar et al.~\cite{FCC24}, Marlin achieves close to optimal 4x improvement over the 16 bit GEMM calculation for batches of up to 16-32 inputs, which translates to
substantial end-to-end inference speedups.
Originally developed for the Nvidia's Ampere architecture and integrated with the GPTQ technique, those kernels have been later extended
to support AWQ as well as paved the way to implementing similarly efficient kernels (Machete~\cite{Wil24}) developed for the more recent Nvidia's Hopper architecture.

\section{Motivation}
We first evaluate the accuracy of RTN with the target precision of 4 and 8 bits (which we refer to as RTN-8 and RTN-4, respectively), and compare it to the accuracy of other, more advanced quantization techniques.
We conduct our experiments using the vLLM framework~\cite{KLZ23} v0.6.4, which we patch to support RTN\footnote{We intend to open-source all our changes to vLLM.}.
For all our experiments, we use one system with 8 Nvidia H100 GPUs with 80GB GPU memory, except for when a model does not fit into memory (e.g., the original Llama3.1-405B model for which we use two such systems).
All experiments are run with Python 3.11, PyTorch 2.5.1 and CUDA 12.6.

We experiment with Llama models~\cite{Llama3.1} of various sizes as well as with other open-source models, such as Phi, Dolly and Mistral, in order to validate our findings beyond Llama models.
We use the \textbf{wikitext} dataset~\cite{MXB16} to measure perplexity and common zero-shot tasks, such as WinoGrande~\cite{SBB19}, ARC~\cite{CCE18}, 
TruthfullQA~\cite{LHE22}, CommonSenseQA~\cite{THL19} and PubMedQA~\cite{JDZ19}, to measure the accuracy of large language models.
For all the experiments, we utilize the open-source \textbf{lm\_eval} toolkit~\cite{lm-eval} with its default settings unless specified otherwise.

We compare the performance of baseline versions (that utilize BFloat16 or FP16 data format) to the one produced with RTN.
We also include results for models quantized with state-of-the-art AWQ~\cite{LTT24} and GPTQ~\cite{FAH22} quantization methods that use INT4 as their target precision
(those models are available on HuggingFace model hub~\cite{hf-modelhub}) as well as for models quantized on-the-fly with BitsAndBytes~\cite{DLB22} (BnB) to INT4.

\newlength{\width}
\width1mm

\begin{table}[!t]
\centering 
\begin{subtable}{1\columnwidth}
\resizebox{\columnwidth}{!}{%
\begin{tabular}{c c c c | c } 
\Xhline{4\arrayrulewidth} 
Llama-3.1 & \#Bits & Method & Avg. (6 tasks) $\uparrow$ & Wikitext $\downarrow$\\
\toprule

\multirow{5}{\width}{8B} & 16 & Baseline & 66.07 & 8.64\\
\cmidrule{2-5}
& 8 & RTN & 66.13 & 8.65\\
\cmidrule{2-5}
& 4 & RTN & 64.32 & 9.57\\
& 4 & GPTQ & 65.07 & 9.04\\
& 4 & AWQ  & 65.30 & 9.10\\
& 4 & BnB  & 65.59 & 9.15\\
\Xhline{2\arrayrulewidth}
\multirow{5}{\width}{70B} & 16 & Baseline & 71.67 & 3.94\\
\cmidrule{2-5}
& 8 & RTN & 71.61 & 3.95\\
\cmidrule{2-5}
& 4 & RTN & 70.46 & 8.19\\
& 4 & GPTQ & 70.89 & 4.61\\
& 4 & AWQ & 71.54 & 4.58\\
& 4 & BnB  & 70.98 & 7.17\\
\Xhline{2\arrayrulewidth}
\multirow{5}{\width}{405B} & 16 & Baseline & 73.99 & 1.74\\
\cmidrule{2-5}
& 8 & RTN & 74.12 & 1.75\\
\cmidrule{2-5}
& 4 & RTN & 73.88 & 2.08\\
& 4 & GPTQ & 73.31 & 1.99\\
& 4 & AWQ & 73.81 & 2.00\\
& 4 & BnB  & 73.84 & 1.95\\
\Xhline{4\arrayrulewidth}
\end{tabular}
}
\end{subtable}

\bigskip
\begin{subtable}{1\columnwidth}

\resizebox{\columnwidth}{!}{%
\begin{tabular}{c c c c | c } 
\Xhline{4\arrayrulewidth} 
Llama-3.2 & \#Bits & Method & Avg. (6 tasks) $\uparrow$ & Wikitext $\downarrow$\\
\toprule
\multirow{5}{\width}{1B} & 16 & Baseline & 51.07 & 15.82\\
\cmidrule{2-5}
& 8 & RTN & 51.10 & 15.83\\
\cmidrule{2-5}
& 4 & RTN & 47.96 & 19.76\\
& 4 & GPTQ & 46.34 & 17.64\\
& 4 & AWQ & 49.26 & 17.32\\
& 4 & BnB  & 49.01 & 17.01\\
\Xhline{2\arrayrulewidth}
\multirow{5}{\width}{3B} & 16 & Baseline & 59.26 & 12.29\\
\cmidrule{2-5}
& 8 & RTN & 59.24 & 12.31\\
\cmidrule{2-5}
& 4 & RTN & 55.74 & 13.34\\
& 4 & GPTQ & 57.55 & 13.16\\
& 4 & AWQ & 57.40 & 12.69\\
& 4 & BnB  & 58.16 & 12.73\\
\Xhline{4\arrayrulewidth}
\end{tabular}
}
\end{subtable}
\caption{Average zero-shot accuracy on six common tasks (the higher is the better) and perplexity measured on the \textbf{wikitext} dataset (the lower is the better), 
measured with three variants of the Llama3.1 model and two variants of the Llama3.2 model.} 
\label{table:llama-acc-short} 
\vspace{4mm}
\end{table}

\begin{figure*}[t!]
\subfloat[][Llama3.2-1B]{\includegraphics[width=0.33\linewidth]{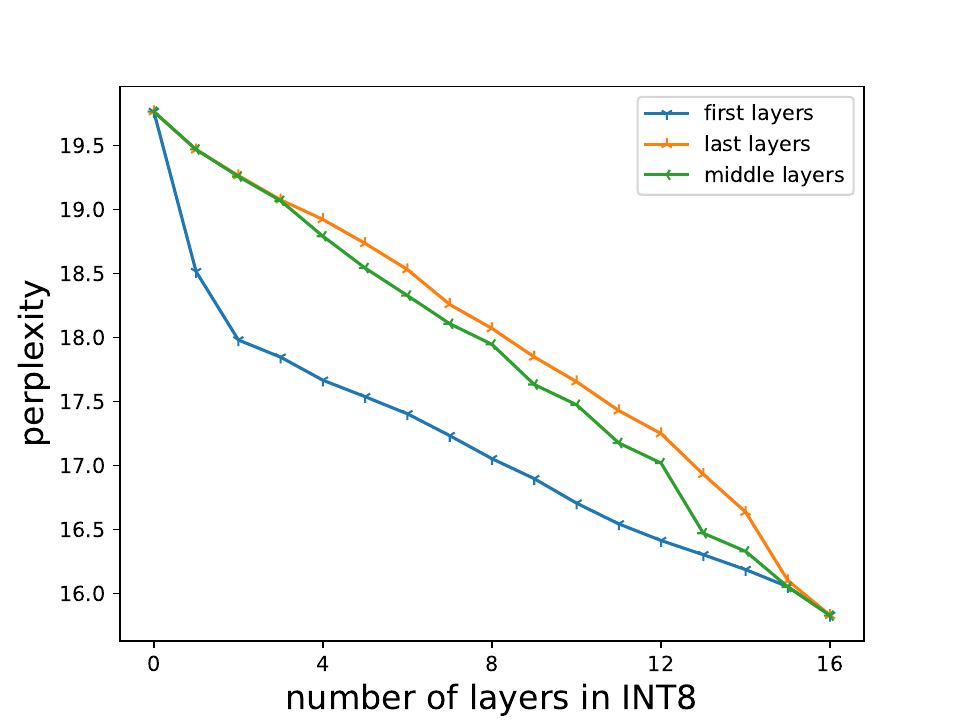}}
\subfloat[][Llama3.1-8B]{\includegraphics[width=0.33\linewidth]{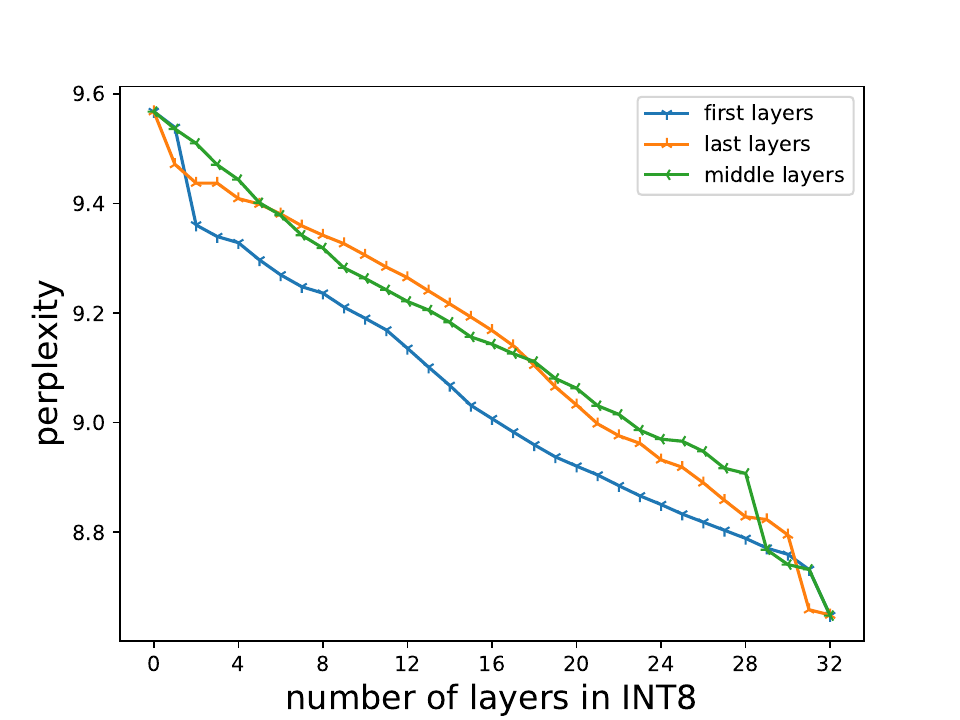}}
\subfloat[][Llama3.3-70B]{\includegraphics[width=0.33\linewidth]{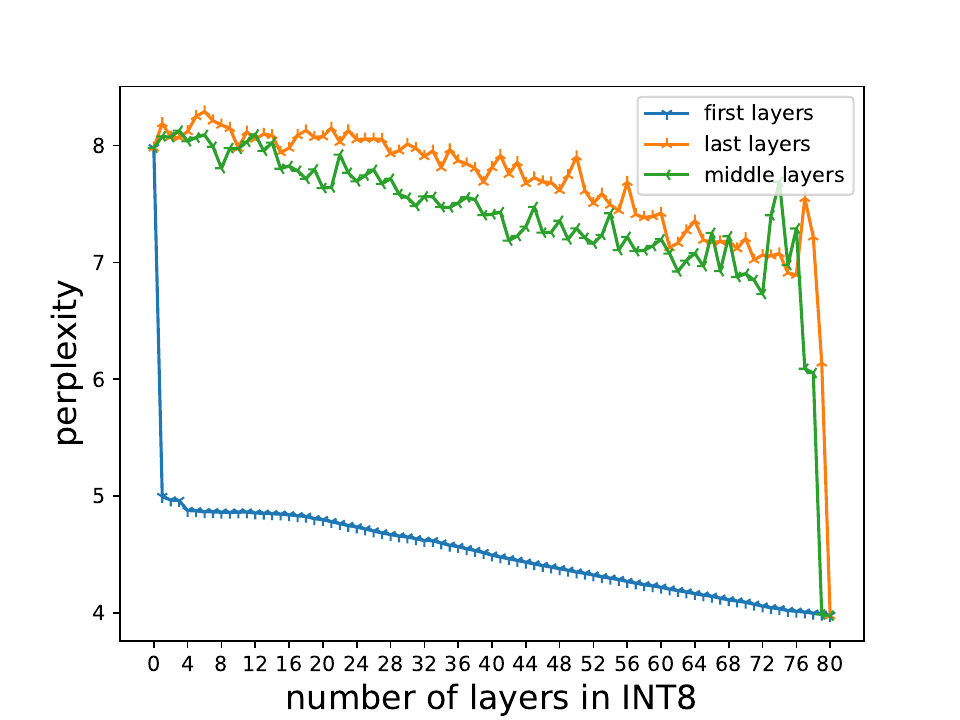}}\\
\subfloat[][Llama3.1-405B]{\includegraphics[width=0.33\linewidth]{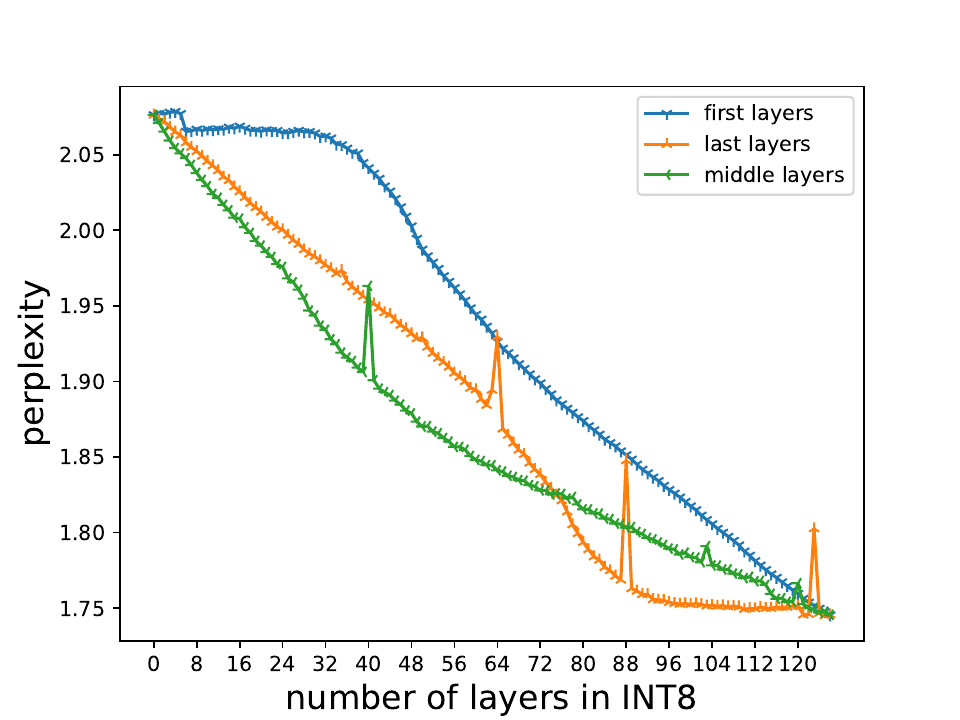}}
\subfloat[][Phi-3-mini-3.8B]{\includegraphics[width=0.33\linewidth]{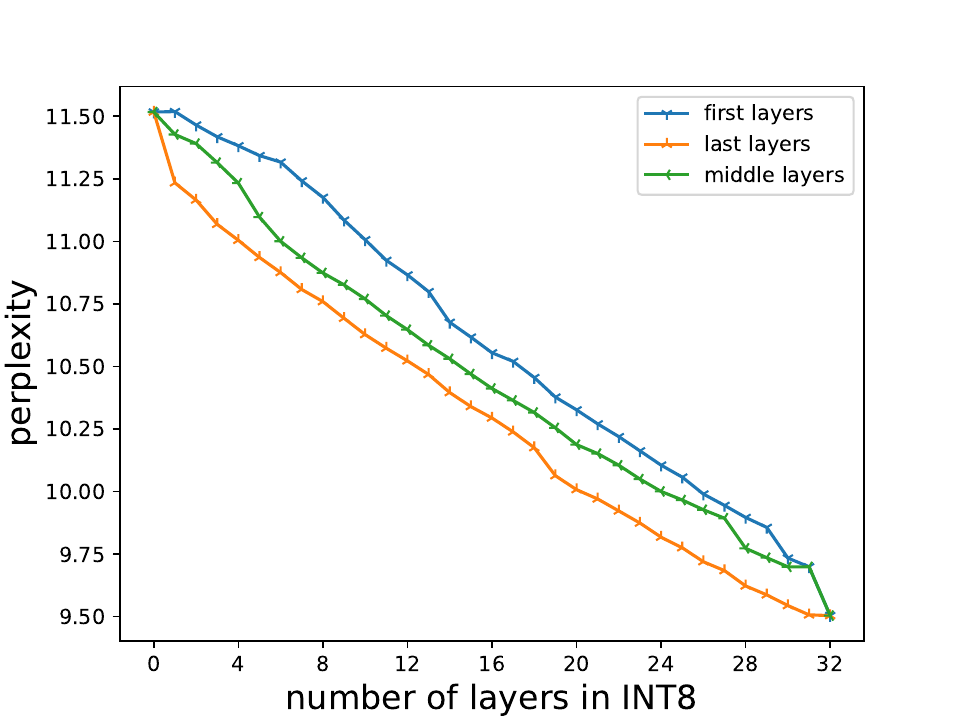}}
\subfloat[][Dolly-v2-3B]{\includegraphics[width=0.33\linewidth]{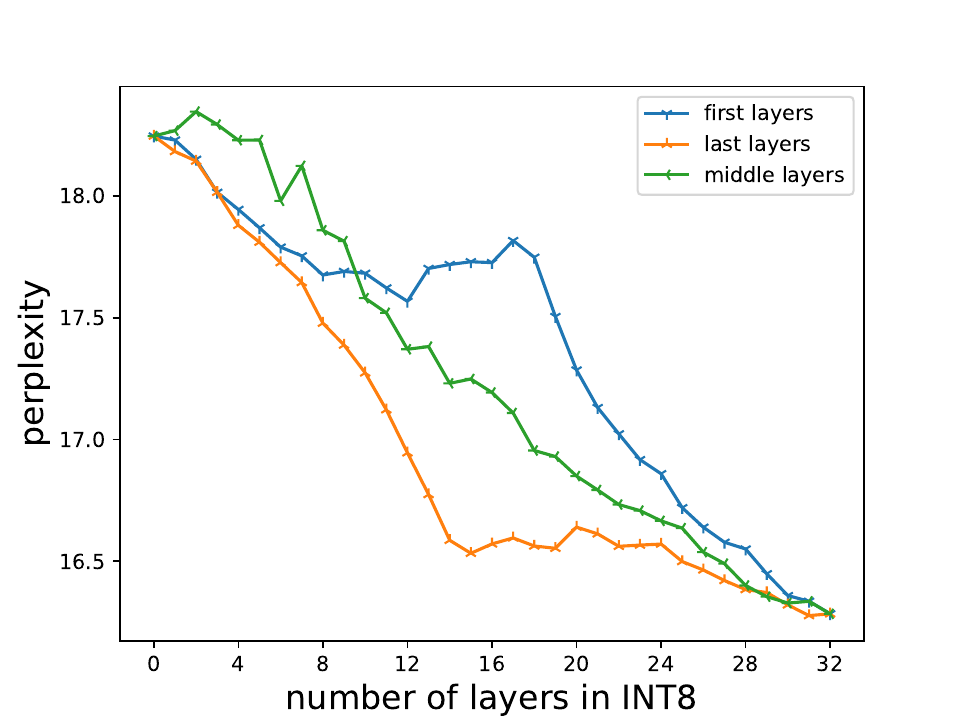}}
\caption{Perplexity of various models on the \textbf{wikitext} dataset (lower is better). Note that the range of the x-axis (number of layers) depends on the model. The y-axis does not start at zero for better readability.}
\label{fig:quantize-x-layers}
\end{figure*}

The results for Llama models are presented in Table~\ref{table:llama-acc-short} (with an extended version available in Appendix~\ref{sec:appendix}).
Two important observations can be made: first, RTN-8 achieves full recovery compared to the baseline versions.
Second, the recovery rate of RTN-4 falls behind those achieved by other quantization techniques, unless the model is extremely large 
(e.g., for Llama 3.1 405B, RTN-4 performs on par with all other variants, including the baseline).
Those findings motivate the following question.
\emph{Can we quantize only a part of the given model to 8 bit (e.g., the first few layers) and keep the rest in 4 bit precision to achieve full accuracy recovery?}
While some theoretical and practical evidence exists that applying quantization non-uniformly across different parts of a model is a viable approach~\cite{ZMC18, ZZL18, HH25}, 
we are not aware of any systematic study evaluating LLM accuracy across variety of tasks.

\section{Selective quantization}
There are two dimensions in the Transformer architecture along which the quantization can be considered.
First, there is the stack of Transformer layers, and so we can select which layers (with all their associated weights) would be quantized into higher precision; 
we call it \emph{horizontal selection}.
Second, each layer typically consists of four linear modules (that contain weight parameters), two in the attention component 
(one of them representing the linear projections of query, key and value, aka QKV)
and two in the fully-connected feed-forward component.
Each of those modules across all Transformer layers can be selected to be quantized to a higher precision; 
we call it \emph{vertical selection}.
In the following subsections, we discuss the empirical findings of selective quantization across those two dimensions.

\subsection{Horizontal selection}
Which layers should be kept in higher precision?
We experiment with three strategies that select either X first, last or middle layers of the Transformer stack to be left in higher precision (i.e., quantized to 8bit)
while the rest of the model is quantized into 4 bits.
We measure the perplexity of the resulting quantized model on the \textbf{wikitext} dataset as we vary X from 1 to the number of layers in a model minus 1.

The results in Figure~\ref{fig:quantize-x-layers} suggest that the model accuracy, as measured by perplexity, is highly sensitive to the 
way Transformer layers are selected.
While for smaller Llama models (Figure~\ref{fig:quantize-x-layers}a--c) the perplexity improves the fastest when the first layers are kept in a higher precision,
for the largest Llama model (Figure~\ref{fig:quantize-x-layers}d) selecting middle layers for higher precision is beneficial most of the time, and yet for
two non-Llama models (Figure~\ref{fig:quantize-x-layers}e--f) it appears the best strategy is to keep the last layers in higher precision.
The result for Llama3.3-70B is especially noteworthy -- the perplexity drops substantially after only one layer is kept in 8 bits.

\begin{figure*}[t!]
\subfloat[][Llama3.2-1B]{\includegraphics[width=0.33\linewidth]{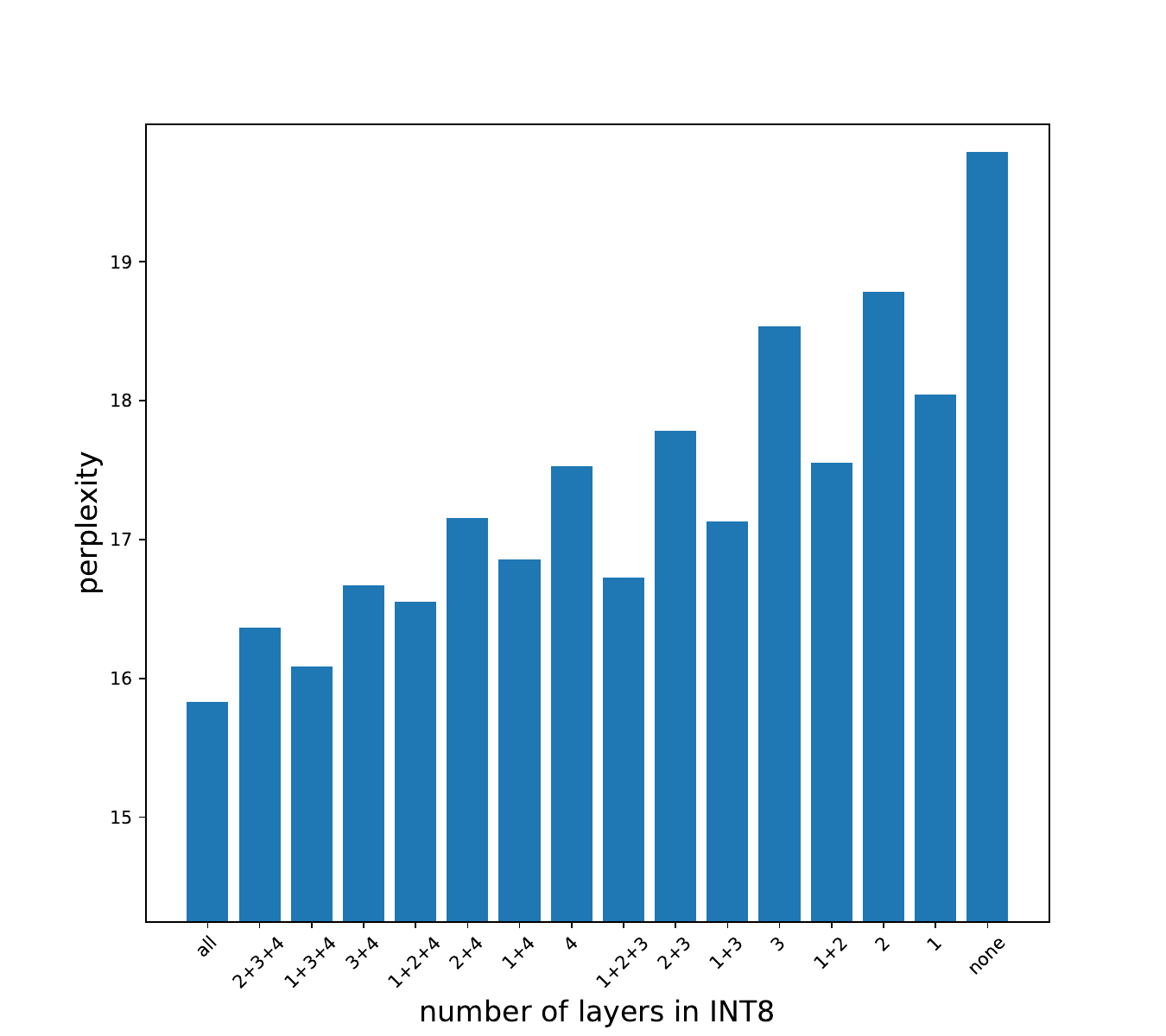}}
\subfloat[][Llama3.1-8B]{\includegraphics[width=0.33\linewidth]{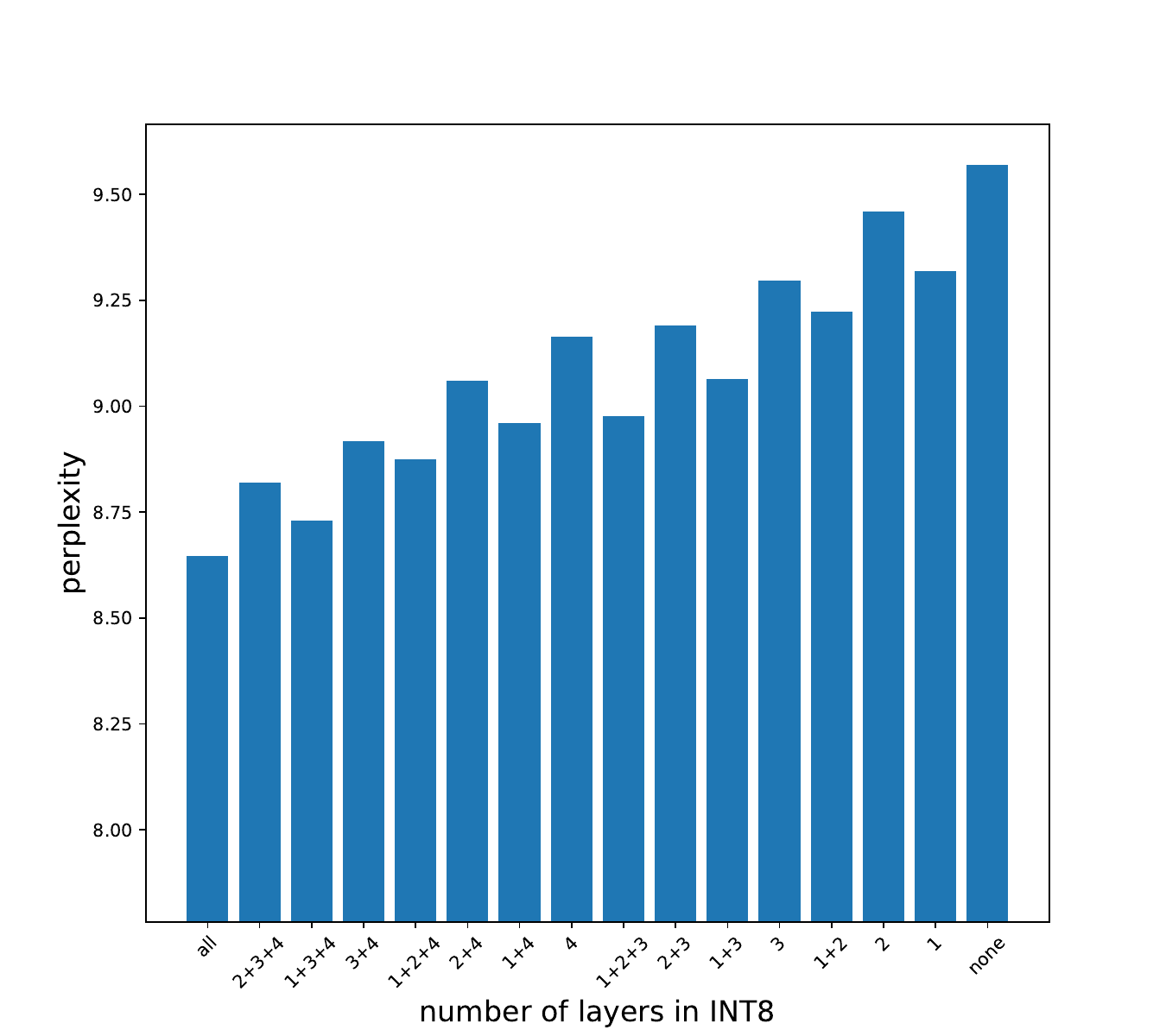}}
\subfloat[][Llama3.3-70B]{\includegraphics[width=0.33\linewidth]{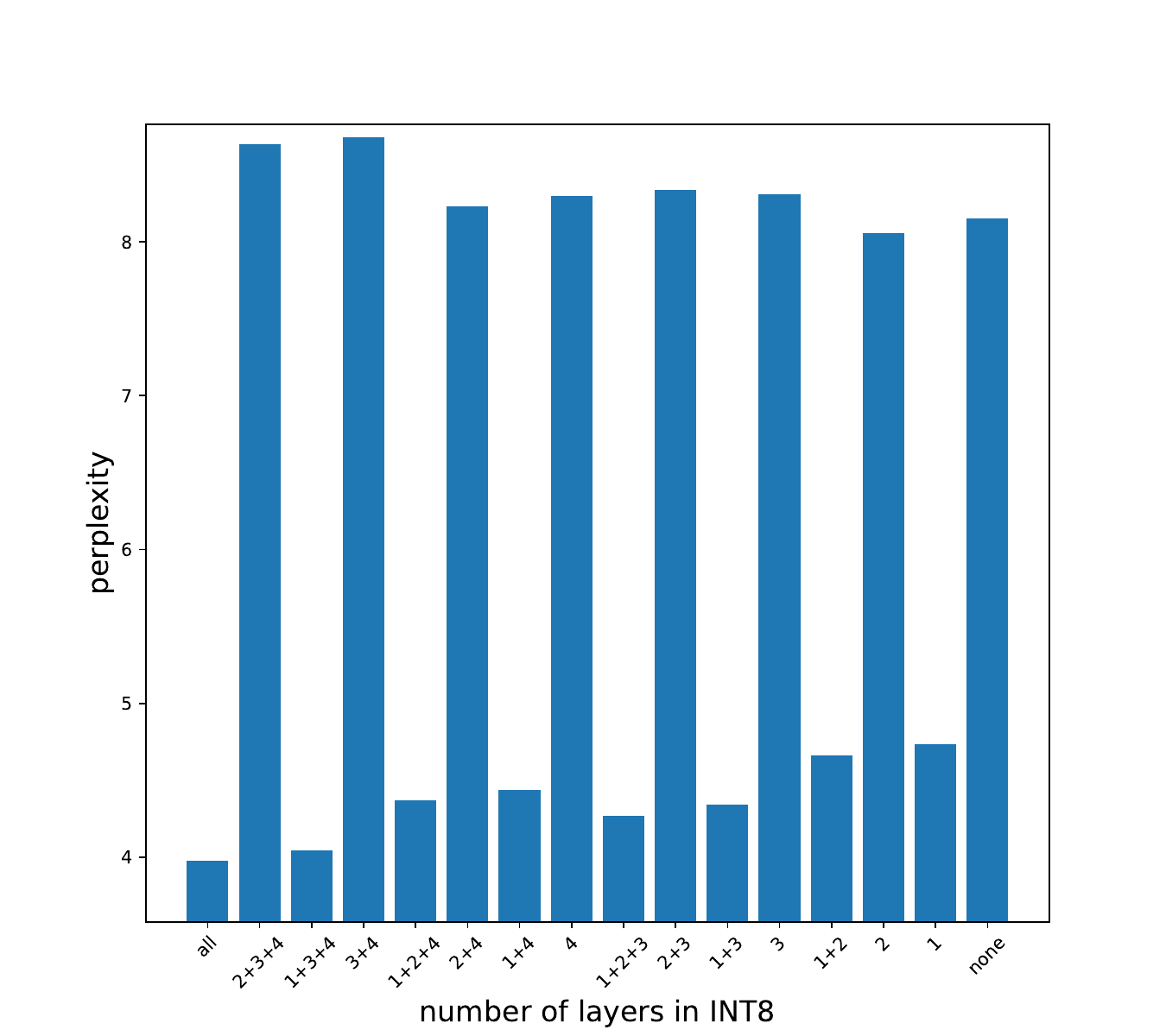}}\\
\subfloat[][Llama3.1-405B]{\includegraphics[width=0.33\linewidth]{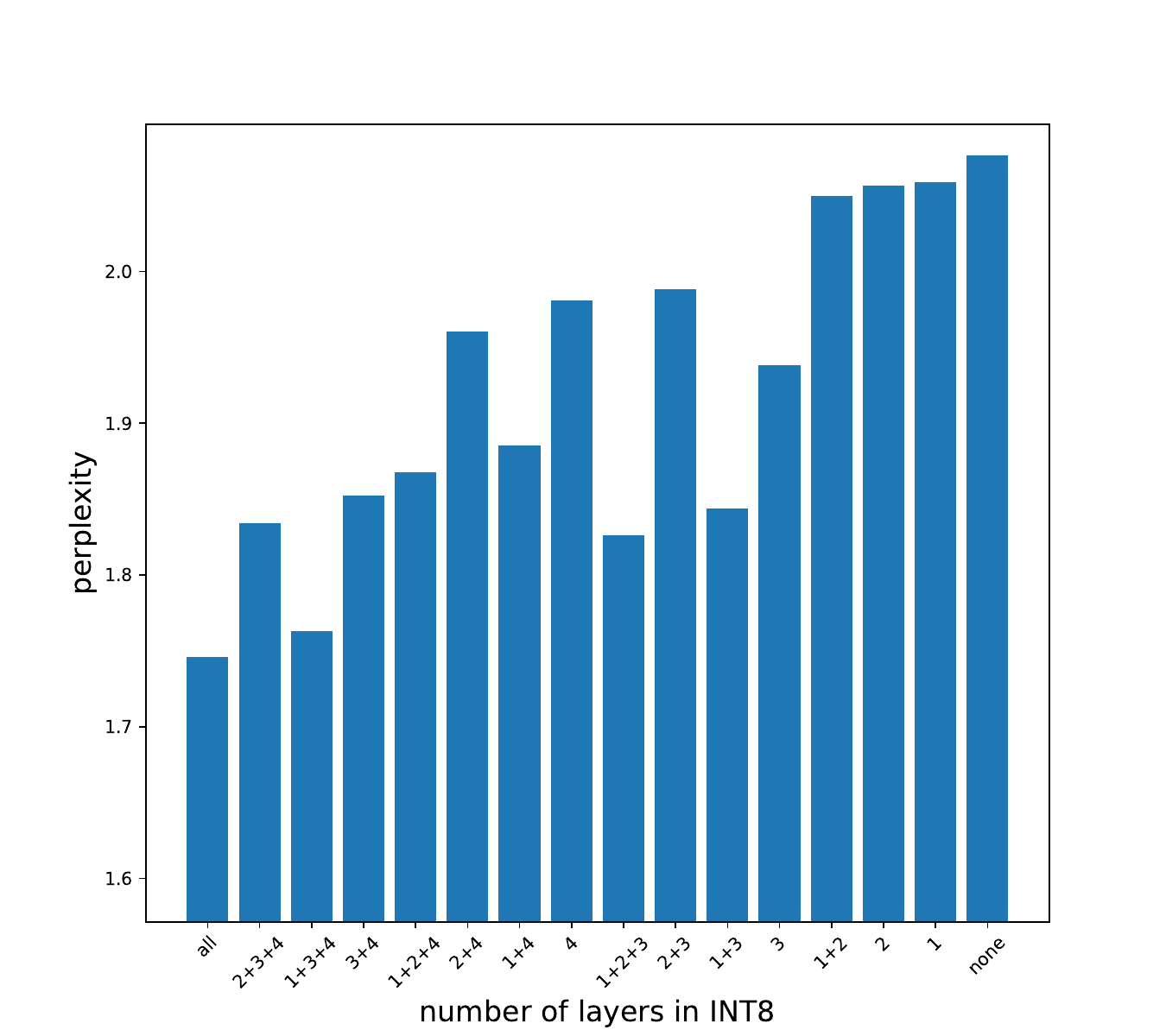}}
\subfloat[][Phi-3-mini-3.8B]{\includegraphics[width=0.33\linewidth]{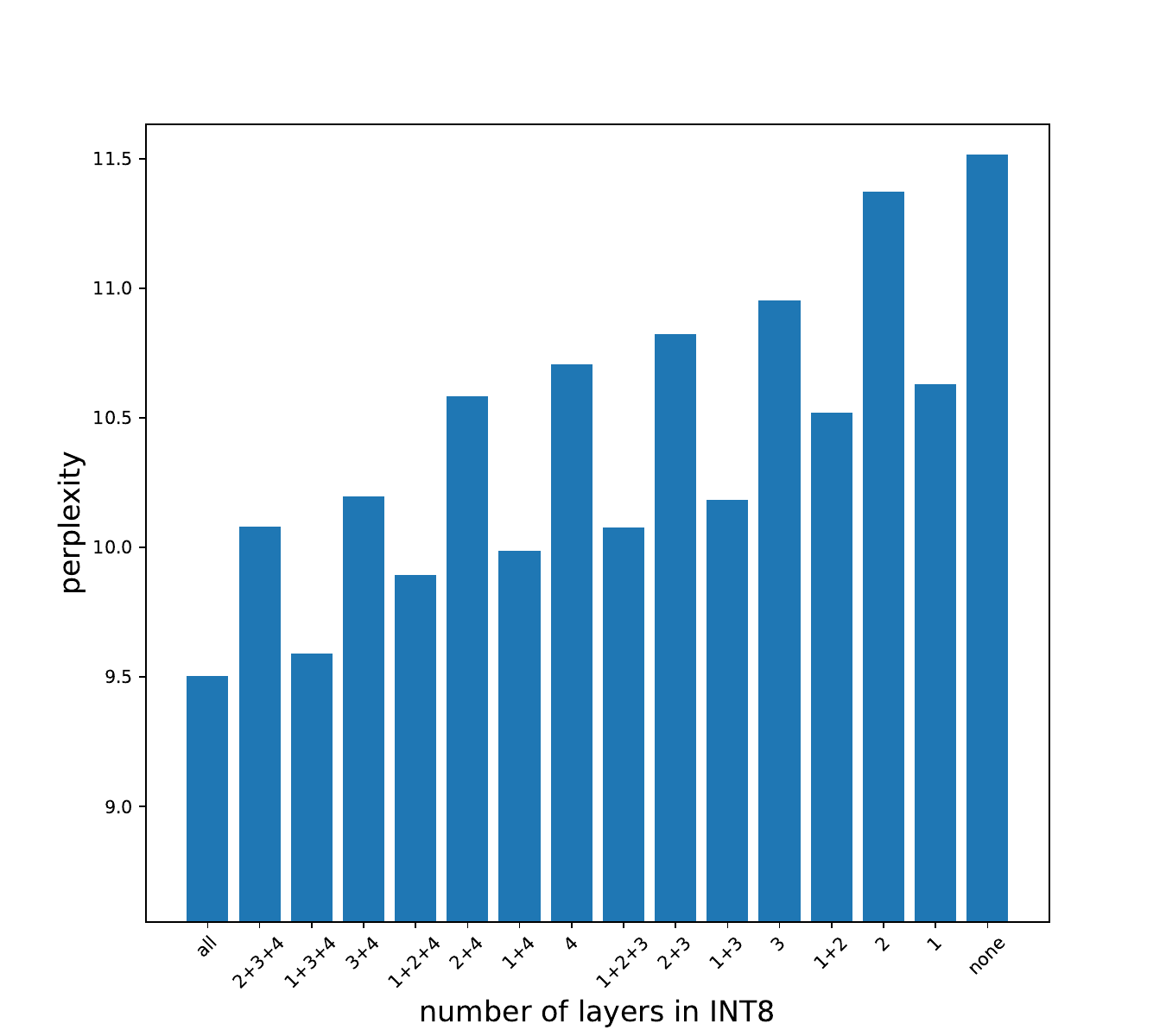}}
\subfloat[][Dolly-v2-3B]{\includegraphics[width=0.33\linewidth]{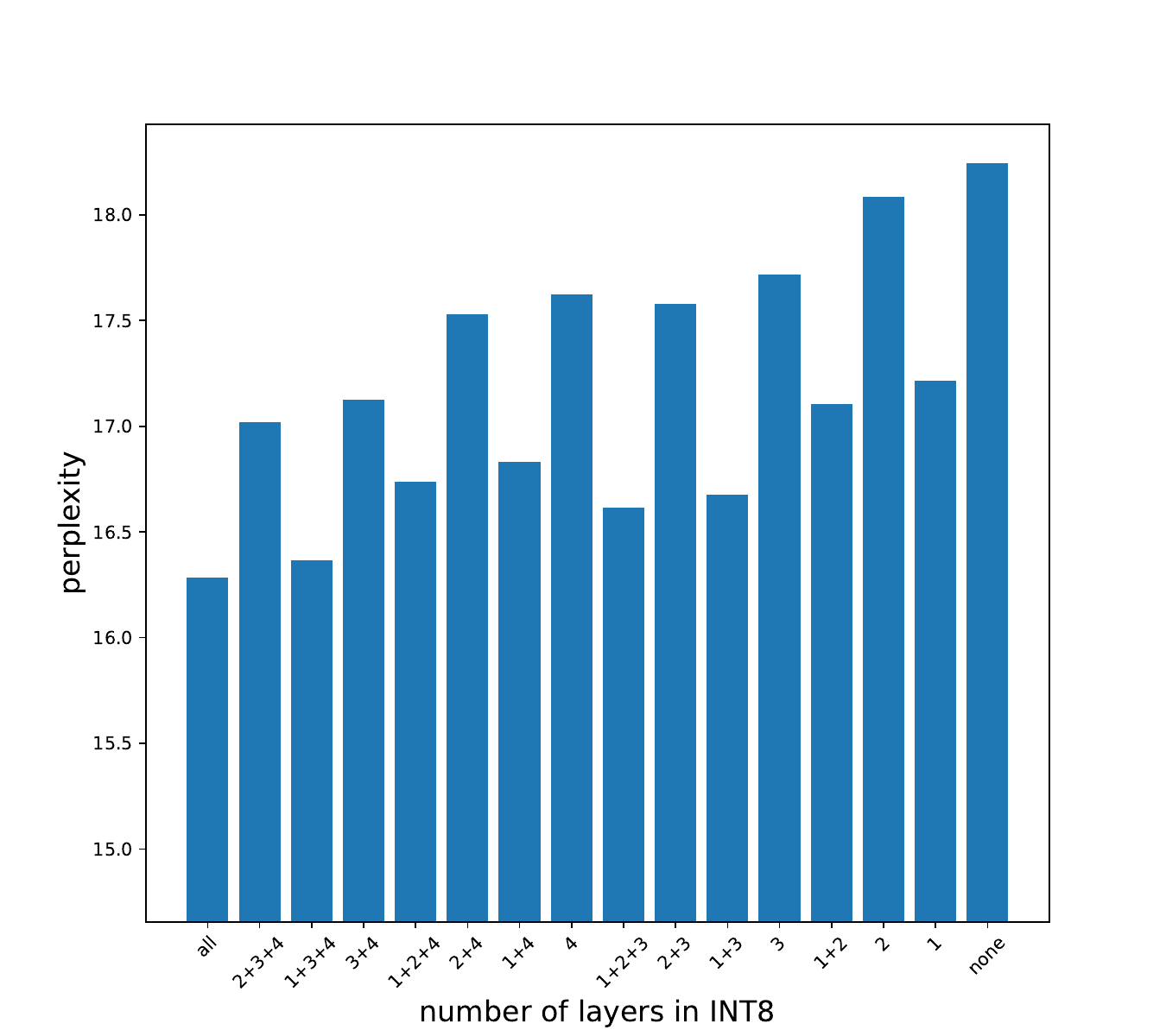}}
\caption{Perplexity of various models on the \textbf{wikitext} dataset (lower is better). The y-axis does not start at zero for better readability.}
\label{fig:quantize-v-modules}
\end{figure*}

\subsection{Vertical selection}
In this section we consider how the accuracy of models depends on which linear modules are selected to be left in higher precision.
Given four linear modules comprising each Transformer layer, we have $2^4 = 16$ ways to make a vertical selection.
In the following charts, we refer to the two modules in the attention component as '1' and '2' and, respectively, to 
the two modules in the fully-connected feed-forward component as '3' and '4'. 
We use the '+' sign for the composition of modules, e.g.,
'1+2+4' designates a selection mode that includes modules '1', '2', and '4'.

The perplexity results for various models and all possible selection modes can be found in Figure~\ref{fig:quantize-v-modules}.
Overall, the perplexity rises as less modules are selected to be left in higher precision.
Notably, however, the perplexity is barely changed when only module '2' (the final projection in the self-attention component) 
is quantized to 4 bits, comparing to the result when all modules kept in 8 bits.
In other words, this experiment suggests that model accuracy is not affected substantially when one specific module is quantized to 4 bits across all layers.

\subsection{Hybrid selection}
Following the observations above, we experiment with a hybrid selections strategy, in which
we select X layers (first for all models but Llama3.1-405B and middle for Llama3.1-405B, based on the results in Figure~\ref{fig:quantize-x-layers}) 
and modules '1+3+4' in those selected layers 
are quantized to 8 bit, while the rest of the model is quantized into 4 bit.
We choose X to be $1/4$ or $1/2$ of the number of the layers.
Note that when $1/4$ of the model uses 8 bit precision while the rest uses 4 bit, 
the model is effectively quantized into 5 bit, on average.
Similarly, for the case of half and half, the model is effectively quantized into 6 bit.
However, since we also quantize one of the four linear modules into 4 bits, the average precision is even lower than that.
Yet, for simplicity sake, we refer to those variants RTN-\~{5} and RTN-\~{6}, respectively.
In addition, for the LLama-3.1 70B model we experiment with a variant in which only modules '1+3+4' of the first layer are
quantized to $8$ bit; the rest of the layers are quantized to 4 bit.
We refer to this variant as RTN-\~{4}.

The results for models RTN-\~{4}, RTN-\~{5} and RTN-\~{6} are presented in Table~\ref{table:llama-sim}.
As expected, the accuracy of the models increases with the increased precision.
In particular, RTN-\~{6} is able to recover most accuracy lost by RTN-4.
The recovery rate of RTN-\~{5} improves over RTN-4, and although it falls slightly below that of RTN-\~{6},
the accuracy of RTN-\~{5} matches or exceeds performance of GPTQ, AWQ and BnB for all considered models.
This suggests that RTN-\~{5} is able to 
strike good tradeoff point between the size of the resulting model and its accuracy.

\remove{
Given that the recovery of RTN-\~{6} (RTN-\~{5}) is similar to that of RTN-6 (RTN-5, respectively), one may wonder whether the non-uniform quantization is really necessary.
We point out two main advantages for the latter.
First, the non-uniform quantization allows much more fine-grained tradeoff between the model size and the resulting accuracy. 
For instance, the effective precision of RTN-\~(4) is
$\frac{4(n1-1) + 8}{n}=4 + 4/n$ bits, where $n$ is the number of layers in the model (e.g., 4.125 bit for LLama3.1-8B model with its 32 layers
or 4.05 for LLama3.1-70B with its 80 layers).
Second, most high-performance CUDA kernels for applying GEMM in low precision assume weights are given in 4 or 8 bit precision.
This includes Marlin kernels, which we base our RTN implementation on and describe in the next section.
The non-uniform quantization allows using those kernels for GEMM instead of the far less efficient approach of dequantizing weights into 16 bit first and 
only then applying GEMM that would normally be required for uniform quantization into 5 or 6 bits.
}

\begin{table}
\centering 
\begin{subtable}{1\columnwidth}
\resizebox{\columnwidth}{!}{%
\begin{tabular}{c c c c | c } 
\Xhline{4\arrayrulewidth} 
Llama-3.1 & Method & Avg. (6 tasks) $\uparrow$ & Wikitext $\downarrow$\\
\toprule

\multirow{4}{\width}{8B} & Baseline &  66.07 & 8.64\\
\cmidrule{2-4}
& RTN-\~{6} &  65.80 & 9.07\\
 & RTN-\~{5} & 64.93 & 9.25\\
\Xhline{2\arrayrulewidth}
\multirow{4}{\width}{70B}  & Baseline &  71.67 & 3.94\\
\cmidrule{2-4}
& RTN-\~{6} & 71.27 & 4.47\\
& RTN-\~{5} &  71.32 & 4.73\\
& RTN-\~{4} & 71.55 & 4.91\\
\Xhline{2\arrayrulewidth}
\multirow{4}{\width}{405B} & Baseline &  73.99 & 1.74\\
\cmidrule{2-4}
& RTN-\~{6} & 74.24 & 1.86\\ 
& RTN-\~{5} &  73.81 & 1.95\\
\Xhline{4\arrayrulewidth}
\end{tabular}
}
\end{subtable}

\bigskip
\begin{subtable}{1\columnwidth}

\resizebox{\columnwidth}{!}{%
\begin{tabular}{c c c c | c } 
\Xhline{4\arrayrulewidth} 
Llama-3.2 & Method & Avg. (6 tasks) $\uparrow$ & Wikitext $\downarrow$\\
\toprule
\multirow{4}{\width}{1B} & Baseline &  51.07 & 15.82\\
\cmidrule{2-4}
& RTN-\~{6} & 50.77 & 17.21\\
& RTN-\~{5} &  50.02 & 17.73\\
\Xhline{2\arrayrulewidth}
\multirow{4}{\width}{3B} & Baseline &  59.26 & 12.29\\
\cmidrule{2-4}
& RTN-\~{6} &  58.46 & 12.91\\
& RTN-\~{5} & 57.54 & 13.13\\
\Xhline{4\arrayrulewidth}
\end{tabular}
}
\end{subtable}
\caption{Zero-shot accuracy on six common tasks (the higher is the better) and perplexity measured on the \textbf{wikitext} dataset (the lower is the better), 
measured with three variants of the Llama3.1 model and two variants of the Llama3.2 model.} 
\label{table:llama-sim} 
\vspace{4mm}
\end{table}

\section{Improving Latency with Marlin}

We base our custom CUDA kernels for RTN on the Marlin project~\cite{FCC24} that implements high performance mixed-precision GEMM operations.
The Marlin kernel expects a specific interleaved data layout for the tensor representing quantized weights.
As a result, we reshuffle the data quantized with RTN to match that specific layout.
This operation takes place once (for each weights tensor), 
while loading and quantizing data, and does not have any impact on the runtime performance during inference.

\begin{figure}[t!]
\includegraphics[width=1.0\linewidth]{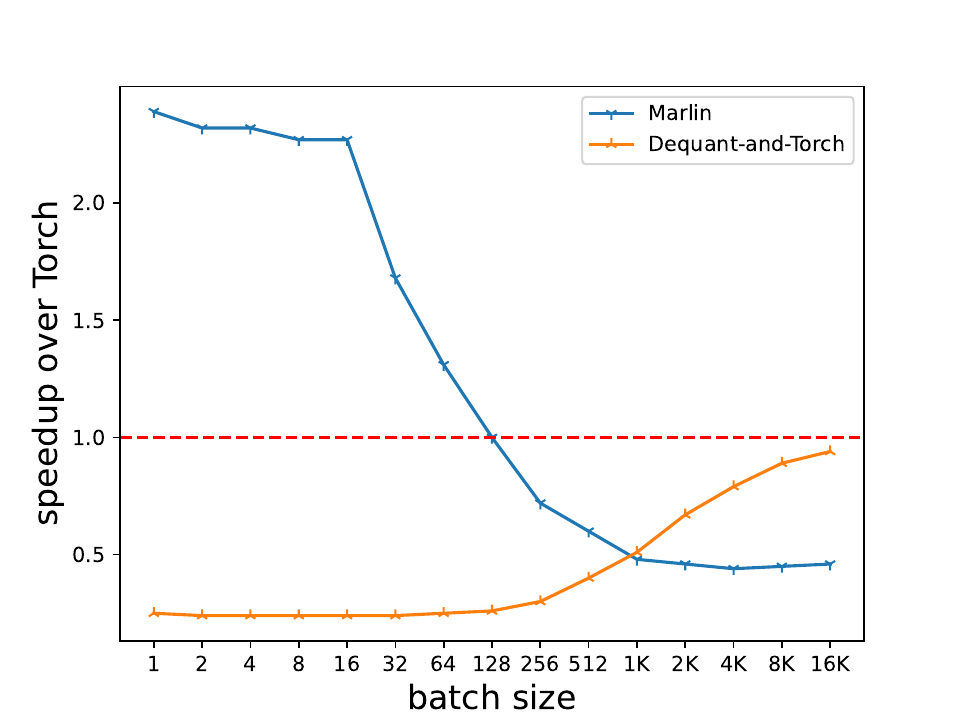}
\caption{Speedup over the default GEMM implementation in PyTorch for tensor shapes corresponding to linear modules of the LLama3.1-8B model}
\label{fig:speedup}
\vspace{4mm}
\end{figure}

Being designed specifically for batch sizes of 16--32 tokens, we discover that the Marlin kernels perform poorly
when they are invoked with tensors representing activations of a large input dimension.
Such tensors are common in realistic workloads since the input dimension is defined by a product of the batch size by the length of (number of tokens in) each input.
Hence, when the batch size is low (or even just 1), a long prompt can lead to activations tensor with a large input dimension.
The issue is highlighted by the data collected with a built-in benchmark in the Marlin repo\footnote{\url{https://github.com/IST-DASLab/marlin/blob/master/bench.py}}.
This benchmark varies the batch size and measures the average latency of running GEMM operations 
with matrices corresponding to shapes employed by linear modules of a  specific model; the results for the LLama3.1-8B model are depicted in Figure~\ref{fig:speedup} where we show the speedup of the Marlin kernel relative to the GEMM operation implemented by PyTorch; 
the speedup falls below $1$ for batch size over $128$ tokens\footnote{We note that the authors of Marlin mention that their kernels target Nvidia's Ampere architecture~\cite{FCC24}.
We performed the same experiment on a machine with A100 GPUs, 
and the results were qualitatively similar, though Marlin kept the advantage over PyTorch for longer, with 
the speedup falling below $1$ for batches over $256$ tokens.}.

We mitigate the issue above by introducing a dual-path approach into the implementation of the forward path for linear module of models quantized with RTN
(this path invokes the GEMM operations).
In particular, when the input dimension of the activations tensor is larger or equal than a certain threshold ($1024$, in our case), 
we switch to a path that dequantizes the weights into the full (16 bit) precision and invokes the PyTorch-based GEMM kernel.
As shown in Figure~\ref{fig:speedup}, the overhead of dequantization relative to the latency of the GEMM
operation decreases as the size of the activations tensor grows.
We note that the original implementation of AWQ in vLLM (which is not based on Marlin) takes a similar dual-path approach\footnote{See \url{https://github.com/vllm-project/vllm/blob/main/vllm/model_executor/layers/quantization/awq.py}}.

The practical consequence of the aforementioned discussion is that one needs to pay attention to the input dimension of the activations tensor during inference.
Specifically, by default, vLLM slices the given prompt into chunks of a certain size ($512$ tokens in v0.6.4, changed later to $2048$ tokens in v0.7.0).
As Figure~\ref{fig:speedup} shows, inputs of $512$ tokens lead to nearly the worst slowdown for Marlin kernels over the baseline GEMM implementation.
As a result, we set the prompt chunk size (controlled by the \textbf{max\_num\_batched\_tokens} parameter in vLLM) to $16K$ in all 
latency experiments described below.
We note that this setting improves performance not only for RTN, but for all other evaluated baselines, as it helps to utilize better 
tensor cores involved in executing GEMM by increasing the amount of computation per each GEMM invocation.

We evaluate the generation performance of several Llama models, in its baseline configuration 
as well as when quantized using RTN (with various levels of precision) and using GPTQ and AWQ.
We use two different setups -- one with a short prompt of $256$ tokens and output length of $32$ tokens, and another 
with a long prompt of $4096$ tokens and output length of $512$ tokens.
The first setup represents a short question-answer interaction with a chatbot, while the other represents a query with more context, 
such as one that might be generated by a code completion system.
(Since the results were similar for both setups, we include only figures for the first setup, for brevity).
For all experiments, we use the built-in vLLM latency benchmark\footnote{\url{https://github.com/vllm-project/vllm/blob/main/benchmarks/benchmark_latency.py}} 
and run all variants using the same hardware
(1/4/8 H100 GPUs for Llama-3.1 8B/70B/405B models, respectively). 
Since the baseline Llama-3.1 405B model does not fit into 8 H100 GPUs, we do not collect results for this configuration.

The generation throughput results are shown in Table~\ref{table:llama3.1-latency}.
For small batches, RTN-8 outperforms the baseline by up to $25\%$, 
while RTN-4 outperforms the baseline by up to $37\%$.
The speedup achieved by all RTN variants over BnB is substantial, especially for small batches.
This result highlights the performance cost paid by BnB for improving accuracy by separating and handling
outliers in model weights. 
Note that RTN-4 also outperforms GPTQ and AWQ (on small batches) despite that all three methods employ
Marlin-based kernels and 4bit precision.
We attribute this advantage of RTN to its relative simplicity that, among other things, does not consider zero 
points for its quantization groups (unlike both GPTQ and AWQ), hence requiring less floating point operations 
for dequantization.
As expected, the performance of RTN-\~{6} and RTN-\~{5} falls in between RTN-8 and RTN-4.
For large batches, RTN variants trail closely the baseline thanks to its dual-path approach discussed above.
Marlin-based kernels for AWQ and GPTQ do not utilize this approach and hence perform poorly in this case.

\setlength{\tabcolsep}{1.5pt}
\begin{table}[!t]
\centering 
\resizebox{\columnwidth}{!}{%
\begin{tabular}{c c | c c c c c c c c |} 
\Xhline{4\arrayrulewidth} 
& & \multicolumn{8}{c|}{Input $256$ / Output $32$} \\
Llama-3.1 & Batch & Baseline & RTN-8 & RTN-\~{6}  & RTN-\~{5}  & RTN-4 & GPTQ & AWQ & BnB \\ 
\toprule
\multirow{7}{\width}{8B} & 1 & 8.03 & 6.05 & 5.47 & 5.2 & 5.06 & 6.31 & 5.4 & 9.36\\
& 4 & 2.1 & 1.71 & 1.58 & 1.51 & 1.47 & 1.71 & 1.63 & 7.08\\
& 16 & 0.7 & 0.61 & 0.56 & 0.57 & 0.55 & 0.63 & 0.74 & 1.98\\
& 64 & 0.36 & 0.35 & 0.35 & 0.35 & 0.35 & 0.38 & 0.58 & 0.69\\
& 256 & 0.29 & 0.31 & 0.31 & 0.31 & 0.31 & 0.32 & 0.53 & 0.4\\
& 1024 & 0.29 & 0.31 & 0.31 & 0.31 & 0.31 & 0.32 & 0.53 & 0.4\\
\Xhline{2\arrayrulewidth}
\multirow{7}{\width}{70B} & 1 & 19.38 & 14.86 & 13.31 & 12.73 & 12.13 & 18.32 & 13.22 & 25.24\\
& 4 & 5.3 & 4.3 & 3.93 & 3.77 & 3.65 & 5.39 & 4.09 & 17.57\\
& 16 & 1.79 & 1.53 & 1.44 & 1.4 & 1.36 & 2.26 & 1.89 & 4.88\\
& 64 & 0.92 & 0.92 & 0.91 & 0.91 & 0.9 & 1.55 & 1.51 & 1.74\\
& 256 & 0.73 & 0.81 & 0.79 & 0.78 & 0.78 & 1.4 & 1.37 & 0.99\\
& 1024 & 0.73 & 0.81 & 0.79 & 0.78 & 0.78 & 1.4 & 1.37 & 0.99\\
\Xhline{2\arrayrulewidth}
\multirow{7}{\width}{405B} & 1 & N/A & 32.59 & 29.22 & 26.87 & 25.23 & 54.17 & 27.7 & 59.91\\
& 4 & N/A & 9.96 & 8.98 & 8.59 & 8.19 & 16.59 & 9.39 & 47.73\\
& 16 & N/A & 3.88 & 3.63 & 3.54 & 3.43 & 7.31 & 4.99 & 13.38\\
& 64 & N/A & 2.36 & 2.33 & 2.32 & 2.32 & 5.2 & 4.08 & 4.72\\
& 256 & N/A & 2.22 & 2.19 & 2.18 & 2.17 & 4.93 & 3.9 & 2.7\\
& 1024 & N/A & 2.22 & 2.19 & 2.18 & 2.16 & 4.92 & 3.9 & 2.7\\
\Xhline{4\arrayrulewidth}
\end{tabular}
}
\caption{Token generation throughput (in msec/token) measured with three variants of the Llama3.1 model, on prompts of 256 tokens and output length of 32 tokens. } 
\label{table:llama3.1-latency} 
\vspace{4mm}
\end{table}

\section{Discussion}
RTN is one of the simplest and easily applicable, data-free quantization techniques for LLMs and beyond.
In this paper we demonstrate how enhancing RTN through selective quantization combined with efficient CUDA kernels can propel it
to match and often surpass more sophisticated techniques that are not data-free.
In particular, selective quantization enables a fine-grained tradeoff between space and accuracy;
in one particular case of the popular open-source Llama-3.1/3.3 70B models, less than $0.05$ extra bits, on average, per each parameter
is all that is needed to close the accuracy gap between RTN and other state-of-the-art quantization techniques.
At the same time, the efficient CUDA kernel allows RTN to achieve higher generation throughput than its more advanced competitors.
This positions RTN as a viable and practical choice for quantizing LLMs.

As more extremely large LLMs adopt the MoE (mixture of experts) architecture~\cite{DeepSeek, Llama4}, 
the selective quantization is well positioned to be applied to such models.
For instance, certain experts (e.g., \emph{generalists}) might be more critical to the end-to-end accuracy of the model than others;
the selective quantization can seamlessly quantize the former to 8 bits and the latter to 4.
Exploring the application of the selective quantization to such models is left for the future work.
Another interesting question we want to consider is automating the selection of layers and modules for higher precision under given
memory budget and/or accuracy degradation target.

\section*{Limitations}
As presented, this work has several limitations, which we aim to address in future work.
First, our evaluation puts heavy focus on Llama models.
Although those models are a popular choice in open-source community, and we do compliment our experiments with other
open-source models, robust evaluation of more models of various sizes would strengthen our findings further.
In particular, as newer models adopt the MoE (mixture of experts) architecture, including the latest generation of Llama 4 models~\cite{Llama4}, 
it would be interesting to confirm the application of our ideas to such models as well.
(We note that our experiments with a Mixtral 8x22B model provide evidence that they apply).

Second, while selective RTN can match and outperform other quantization techniques, the selection of layers and modules for higher precision 
is limited to a manual process. 
Automating this process, all while maintaining the practical advantages of RTN, including its data-free nature, remains an open question.

\section*{Acknowledgements}
The author would like to thank Jingqiao Zhang, Ming Lin and Arthur Cheng for valuable discussions and evaluation of ideas presented in this paper.

\bibliography{rtn}

\appendix

\section{Extended Accuracy Results}
\label{sec:appendix}

\begin{table*}[ht]
\centering 
\begin{subtable}{1\textwidth}
\resizebox{\textwidth}{!}{%
\begin{tabular}{c c c c c c c c c c | c } 
\Xhline{4\arrayrulewidth} 
Llama-3.1 & \#Bits & Method & Winogrande & ARC-e & ARC-c & TruthfulQA & CommonSenseQA & PubMedQA & Avg. & Wikitext\\
\toprule
\multirow{5}{\width}{8B} & 16 & Baseline &73.64 & 81.94 & 51.71 & 36.96 & 77.15 & 75.00 & 66.07 & 8.64\\
\cmidrule{2-11}
& 8 & RTN & 73.72 & 81.78 & 51.79 & 36.84 & 77.07 & 75.60 & 66.13 & 8.65\\
\cmidrule{2-11}
& 4 & RTN & 74.27 & 80.89 & 50.26 & 35.25 & 74.86 & 70.40 & 64.32 & 9.57\\
& 4 & GPTQ & 71.98 & 81.44 & 50.85 & 35.01 & 74.77 & 76.40 & 65.07 & 9.04\\
& 4 & AWQ & 73.09 & 80.89 & 51.02 & 34.76 & 75.84 & 76.20 & 65.30 & 9.10\\
& 4 & BnB & 73.56 & 81.06 & 51.54 & 36.47 & 75.51 & 75.40 & 65.59  & 9.15\\
\Xhline{2\arrayrulewidth}
\multirow{5}{\width}{70B} & 16 & Baseline & 79.01 & 86.78 & 62.54 & 40.64 & 80.67 & 80.40 & 71.67 & 3.94\\
\cmidrule{2-11}
& 8 & RTN & 79.16 & 86.62 & 62.46 & 40.51 & 80.51 & 80.40 & 71.61 & 3.95\\
\cmidrule{2-11}
& 4 & RTN & 77.27 & 85.56 & 60.92 & 40.27 & 80.34 & 78.40 & 70.46 & 8.19\\
& 4 & GPTQ & 78.14 & 86.11 & 61.18 & 40.39 & 79.93 & 79.60 & 70.89 & 4.61\\
& 4 & AWQ & 79.40 & 86.62 & 62.12 & 40.27 & 80.84 & 80.00 & 71.54 & 4.58\\
& 4 & BnB & 79.24 & 85.52 & 60.84 & 40.88 & 80.18 & 79.20 & 70.98 & 7.17\\
\Xhline{2\arrayrulewidth}
\multirow{5}{\width}{405B} & 16 & Baseline & 82.64 & 88.38 & 64.76 & 46.27 & 84.68 & 77.20 & 73.99 & 1.74\\
\cmidrule{2-11}
& 8 & RTN & 82.79 & 88.17 & 64.85 & 46.51 & 84.77 & 77.60 & 74.12 & 1.75\\
\cmidrule{2-11}
& 4 & RTN & 83.03 & 88.26 & 64.76 & 45.65 & 84.36 & 77.20 & 73.88 & 2.08\\
& 4 & GPTQ & 81.37 & 88.26 & 65.10 & 46.14 & 83.62 & 75.40 & 73.31 & 1.99\\
& 4 & AWQ & 82.72 & 88.26 & 64.59 & 45.90 & 84.19 & 77.20 & 73.81 & 2.00\\
& 4 & BnB & 82.79 & 88.17 & 64.33 & 47.12 & 84.44 & 76.20 & 73.84 & 1.95\\
\Xhline{4\arrayrulewidth}
\end{tabular}
}
\end{subtable}
\bigskip
\begin{subtable}{1\textwidth}
\resizebox{\textwidth}{!}{%
\begin{tabular}{c c c c c c c c c c | c } 
\Xhline{4\arrayrulewidth} 
Llama-3.2 & \#Bits & Method & Winogrande & ARC-e & ARC-c & TruthfulQA & CommonSenseQA & PubMedQA & Avg. & Wikitext\\
\toprule
\multirow{5}{\width}{1B} & 16 & Baseline &  59.67 & 68.73 & 35.58 & 27.05 & 55.36 & 60.00 & 51.07 & 15.82\\
\cmidrule{2-11}
& 8 & RTN & 59.43 & 68.35 & 35.41 & 27.17 & 55.45 & 60.80 & 51.10 & 15.83\\
\cmidrule{2-11}
& 4 & RTN &  59.12 & 63.34 & 32.17 & 24.85 & 48.48 & 59.80 & 47.96 & 19.76\\
& 4 & GPTQ & 57.46 & 63.51 & 31.83 & 26.07 & 41.20 & 58.00 & 46.34 & 17.64\\
& 4 & AWQ & 59.19 & 66.67 & 34.73 & 25.46 & 51.52 & 58.00 & 49.26 & 17.32\\
& 4 & BnB & 57.38 & 65.32 & 35.32 & 25.46 & 53.15 & 57.40 & 49.01 & 17.01\\
\Xhline{2\arrayrulewidth}
\multirow{5}{\width}{3B} & 16 & Baseline & 67.88 & 74.41 & 43.43 & 32.31 & 67.73 & 69.80 & 59.26 & 12.29\\
\cmidrule{2-11}
& 8 & RTN & 67.32 & 73.91 & 43.60 & 32.68 & 67.73 & 70.20 & 59.24 & 12.31\\
\cmidrule{2-11}
& 4 & RTN & 66.30 & 69.74 & 39.93 & 31.33 & 62.57 & 64.60 & 55.74 & 13.34\\
& 4 & GPTQ & 65.75 & 71.89 & 41.72 & 30.97 & 65.77 & 69.20 & 57.55 & 13.16\\
& 4 & AWQ & 67.56 & 72.98 & 41.30 & 29.62 & 65.36 & 67.60 & 57.40 & 12.69\\
& 4 & BnB & 67.64 & 72.26 & 41.98 & 29.99 & 66.50 & 70.60 & 58.16 & 12.73\\
\Xhline{4\arrayrulewidth}
\end{tabular}
}
\end{subtable}
\caption{Zero-shot accuracy on six common tasks (the higher is the better) and perplexity measured on the \textbf{wikitext} dataset (the lower is the better), measured with three variants of the Llama3.1 model and two variants of the Llama3.2 model.} 
\label{table:llama-acc} 
\end{table*}

\end{document}